\documentclass[conference,a4paper]{IEEEtran}
\IEEEoverridecommandlockouts
\usepackage{algorithm}
\usepackage{amsmath}
\usepackage{amsfonts}
\usepackage{amssymb}
\usepackage{cite}
\usepackage{algpseudocode}
\usepackage{graphicx}
\usepackage{textcomp}
\usepackage{xcolor}
\usepackage{booktabs}
\usepackage{dsfont}
\usepackage{algorithm}
\usepackage{algpseudocode}

\def\BibTeX{{\rm B\kern-.05em{\sc i\kern-.025em b}\kern-.08em
    T\kern-.1667em\lower.7ex\hbox{E}\kern-.125emX}}

\begin{document}
\setlength{\columnsep}{0.21in}

\title{\LARGE Communication-Aware Knowledge Distillation for \\ Federated LLM Fine-Tuning over Wireless Networks}

\author{%
  \IEEEauthorblockN{Xinlu Zhang, Na Yan, Yang Su, Yansha Deng, Toktam Mahmoodi}%
  \IEEEauthorblockA{Department of Engineering, King's College London, London, UK\\
  \{xinlu.zhang, na.2.yan, yang.2.su, yansha.deng, toktam.mahmoodi\}@kcl.ac.uk}%
}

\maketitle

\begin{abstract}
Federated learning (FL) for large language models (LLMs) offers a privacy-preserving scheme, enabling clients to collaboratively fine-tune locally deployed LLMs or smaller language models (SLMs) without exchanging raw data. While parameter-sharing methods in traditional FL models solves number of technical challenges, they still incur high communication overhead and struggle with adapting to heterogeneous model architectures. Federated distillation, a framework for mutual knowledge transfer via shared logits, typically offers lower communication overhead than parameter-sharing methods. However, transmitting logits from LLMs remains challenging for bandwidth-limited clients due to their high dimensionality. In this work, we focus on a federated LLM distillation with efficient communication overhead. To achieve this, we first propose an adaptive Top-k logit selection mechanism, dynamically sparsifying logits according to real-time communication conditions. Then to tackle the dimensional inconsistency introduced by the adaptive sparsification, we design an adaptive logits aggregation scheme, effectively alleviating the artificial and uninformative inputs introduced by conventional zero-padding methods. Finally, to enhance the distillation effect, we incorporate LoRA-adapted hidden-layer projection from LLM into the distillation loss, reducing the communication overhead further while providing richer representation. Experimental results demonstrate that our scheme achieves superior performance compared to baseline methods while effectively reducing communication overhead by approximately 50\%.
\end{abstract}


\section{Introduction}
In the past few years, large language models (LLMs) have achieved remarkable results in natural language processing tasks such as question answering and text generation. To further advance model performance, both academia and industry have invested substantial computational resources and financial capital into scaling model sizes from hundreds of millions to hundreds of billions of parameters. Notable examples of such LLMs include BERT and LLaMA~\cite{zhao2023survey}. However, the centralized training paradigm of LLMs requires uploading large amounts of users' private data, thereby raising significant privacy concerns. Federated learning, a well-established distributed learning paradigm, addresses this issue by allowing the client to collaboratively fine-tune the LLMs locally, avoiding direct exposure of sensitive information. 
The massive size of LLMs leads to frequent model exchanges and voluminous data transfers, resulting in substantial communication overhead and significantly higher uplink bandwidth consumption. Several optimization algorithms were proposed before, but none targeted LLMs\cite{zhang2024joint}. Parameter-efficient fine-tuning (PEFT) offers a straightforward solution to the high communication overhead and fine-tuning cost of federating LLMs. By updating only a small subset of parameters \cite{zhang2024towards}, PEFT markedly reduces the size of updates transmission and their storage burden, thereby lowering on-device computational demand. 

Nonetheless, clients remain constrained by limited uplink bandwidth, which continues to hinder efficient collaborative training across the federation. Moreover, while PEFT reduces the amount of data that needs to be shared between clients and the server, it becomes much harder and more costly to combine updates when clients use different model architectures, making aggregation on the server more complex. On the other hand, modern LLM weights and configurations such as GPT and LLaMA are valuable intellectual property; even transmitting low rank adapters exposes a tangible risk of model reconstruction and information leakage\cite{su2024federated}. Finally, limited uplink bandwidth on mobile or edge devices imposes strict latency constraints that can stall training progress altogether. To alleviate these issues, recent work proposed federated distillation (FedD) for LLMs: to avoid clients downloading full pretrained weights. Instead, they learn from server-side teacher logits on public inputs, which achieves substantial communication savings, preserves model heterogeneity, and reduces privacy risks. Despite this, the scheme maintains competitive performance \cite{yan2025federated}.

However, it is still challenging to efficiently and reliably transfer knowledge from LLMs to clients with limited communication resources in real-world settings. Issues such as noisy logits, bandwidth bottlenecks, and teacher–student distribution mismatches can all undermine performance. To tackle model heterogeneity, early federated-distillation methods such as FedMD \cite{li2019fedmd} and FedDF \cite{lin2020ensemble} dispense with parameter averaging and instead let each client upload its soft predictions, which are then aggregated by the server to train a unified student model. Subsequent variants refined this idea: FedGen\cite{zhu2021data} proposed label-free distillation by leveraging teacher predictions on unlabeled data, while FedKD \cite{wu2022communication} combined adaptive knowledge distillation with dynamic gradient compression for reducing communication overheads. Although these methods are promising for privacy and efficiency, they were mainly designed for relatively small models with fewer parameters and lower computational requirements, and for limited tasks such as image classification or simple NLP benchmarks. As a result, it remains unclear how well these methods can scale to LLMs.Recent research interest has shifted towards FedD across LLMs. However, progress in this area remains in an early stage. FedMKT\cite{fan2024fedmkt},  aligned and fused knowledge among heterogeneous LLM in a federated setting to enhance cross-client collaboration. However, most existing works overlooked the communication overhead introduced by LLMs and the inefficiencies in knowledge transfer caused by diverse model architectures and data distributions. These difficulties highlight the need for more adaptive and robust algorithms, and they motivate our exploration of higher-efficiency, more resilient FedD techniques for LLM.

Efficiently conveying useful information has become a major challenge in FedD for LLM. 
However, when applied to LLMs with rich and multi-level representations, relying solely on output logits is often insufficient to capture the full spectrum of latent feature information. This limitation can lead to degraded task performance.
Moreover, most existing aggregation schemes rely on simplistic strategies such as zero-padding or mean averaging, where missing dimensions are filled with zeros and all contributions are treated equally. As a result, the aggregation process is unable to retain critical client-specific knowledge. This limits its capacity to model the inherent heterogeneity of local data distributions, thereby impeding convergence to the global optimum.

Motivated by the above considerations, this paper introduces a communication-aware federated distillation framework for LLMs over wireless networks. The main contributions of our research are summarized as follows: 
\begin{enumerate}

\item{We propose a federated LLM distillation framework for bandwidth-constrained settings. It incorporates LoRA-based alignment, channel-aware Top-k logit sparsification, and adaptive weighted aggregation((AdaLD), cutting communication and speeding up knowledge exchange among clients.}


\item{We develop an adaptive logits aggregation mechanism that compensates for the sparsity imposed by bandwidth-aware Top-k filtering, thereby enabling distillation to preserve and transfer the model’s critical information more accurately.}

\item{To further enhance distillation efficiency, we improve the distillation efficiency by injecting LoRA‐induced activation residuals into the loss and jointly learning their scaling, delivering stronger guidance with negligible parameter overhead. }

\item{We simulate our proposed scheme against three baseline methods in SLM for client and LLM for server. Compared with the baselines, our scheme reduce communication overheads by about 50\%  while improves model accuracy, demonstrating the effectiveness of our scheme.}
\end{enumerate}

The rest of the paper is organized as follows: Section II presents the system model. Section III introduces a LoRA-based FedD scheme and an adaptive logits aggregation. Section IV outlines the simulation results. Finally, Section V concludes the paper.

\section{System Model}

Consider a Federated learning network consisting of a cloud server $\mathcal{S}$, distributed with a set of $N$ clients, denoted $\mathcal{N}  = \{  1,2,\ldots,N \}$. Servers and users share the same public dataset without real labels $D_p$, with a sample volume $|D_p|$. Each client $n \in N $ is training on a private dataset $D_n$, with a data volume $|D_n|$. Each data sample is represented by a set of input-output pairs $(X_{n,i}, Y_{n,i}) , i \in \{1,2,\dots,|D_n|\} $, where $X_{n,i}$ denotes the $i$-th input data and $Y_{n,i}$ represents the corresponding ground truth for $X_{n,i}$.

The server holds LLM and the clients hold SLM fine-tuned by LoRA $\boldsymbol{\theta_n}$  under the same architectures. The server and the clients aim to jointly improve the performance of LLM and SLM through joint distillation learning without sharing model parameters. We assume that $n$ clients perform the same task with model weights $W$. As shown in Fig. \ref{figure1}, the iterative learning process is introduced as follows:
\begin{figure*}[t!]
  \centering
  \includegraphics[width=0.8\textwidth]{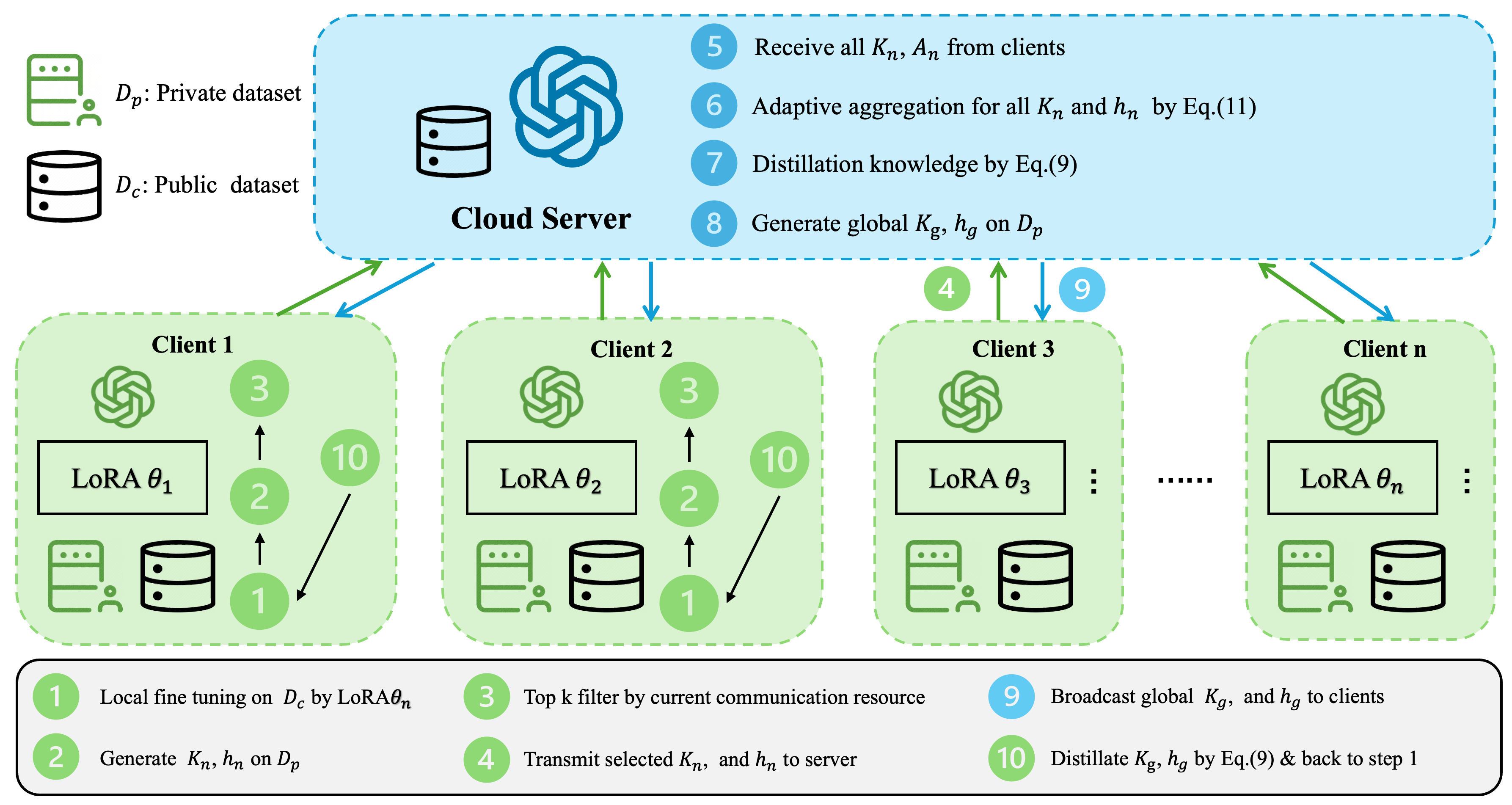}
  \caption{The workflow of AdaLD scheme. Each communication round involves 10 steps to fine-tune the server's LLM and clients’ SLM. }
  \label{figure1}
  \vspace{-7pt}
\end{figure*}

\subsection {Local Fine-tuning and upload } Each client first fine-tunes the model using the LoRA on local private data. Let $W'$ denote the frozen backbone parameters that are shared by all clients, and $\theta_n=\{A_n,B_n\}$ denote the LoRA parameters of client $n$.
$r$ is the rank of the low-rank approximation.\noindent
Here, \(d_{\text{in}}\) and \(d_{\text{out}}\) denote the input and output dimensions of the linear layer. For a linear layer with pretrained weight $W'\!\in\!\mathbb{R}^{d_\text{out}\times d_\text{in}}$, the update of model weights can be defined as
\begingroup
\addtolength{\abovedisplayskip}{0.2em}
\addtolength{\belowdisplayskip}{0.2em}
\begin{equation}
\label{eq:lora}
W_n
\;=\;
\underbrace{W'}_{\text{shared, frozen weight}} \;+\; \underbrace{B_n A_n}_{\text{trainable weight}},  
\end{equation}
\endgroup
where $A_n\in\mathbb{R}^{r\times d_\text{in}}$ projects the input to a
low-rank subspace of dimension $r\!\ll\!d_\text{in}$,
and $B_n\in\mathbb{R}^{d_\text{out}\times r}$ maps it back to the original space \cite{su2024federated}. Therefore, the objective of each client is formulated as 
\begin{equation}
\small \small
\mathcal{L}_n(\theta_n)= \frac{1}{\lvert D_{n}\rvert} \sum_{(x_{n,i},y_{n,i})\in D_{n}} \bigl(f(x_{n,i};\,W',\theta_n),\,y_{n,i}\bigr)
+\lambda\,\mathcal{R}(\theta_n)\,,
\label{eq1}
\end{equation}
where \(\mathcal{R}(\theta_n)\) is an optional regularizer on \(\theta_n\) with weight \(\lambda\). Then, each client $n$ computes the output logits $K_n$ on the public dataset. 

In addition, our proposed scheme requires extracting the intermediate outputs of the LoRA $\theta_n$. Then upload both of these to the server. To accommodate real-time network conditions, each client $n$ adaptively sparsifies its predicted logits via a Top-k selection. The sparse logits are then transmitted.

\subsection {Global Aggregation and Knowledge Distillation}

After receiving the logits from all clients, the server aggregates all sparse logits to generate the global logits $K_g$. To aggregate the knowledge of all $N$ clients, the server performs the adaptive aggregation proposed for the global aggregated logits. The aggregated logits vector $K_g$ can then be normalized using a softmax function to obtain a global soft label distribution. It serves as the global teacher knowledge, which can be utilized by clients for further local distillation updates. The server then computes the distillation loss by comparing these aggregated global logits with the logits produced by its own LLM on the public dataset. In our proposed scheme, we redefine the calculation of the distillation loss and the aggregation of logits. A detailed introduction will be provided in Section III. After that, the cloud server computes the output logits and LoRA Projection on the public dataset and broadcasts them to the clients.

\subsection{Local Distillation on Clients}

After receiving the global logits and LoRA projection, each client performs local knowledge distillation with the same loss as the server, aligning its predictions with global knowledge. These three stages complete one communication round.

\section {Adaptive Logits Aggregation and LoRA Projection Distillation  Scheme}
In this section, to address the performance degradation caused by communication constraints, we subsequently propose a communication-aware adaptive weighted aggregation scheme for logits. To enhance the distillation efficiency of LLMs, we then introduce a distillation scheme based on intermediate projection from the LoRA adapter.

\subsection{Adaptive Logits Global Aggregation}
To minimize the bandwidth required for uploading, client $n$ selects the Top-k values from its predicted logits before sending them to the server. This method means that for each input sample $x$, only the first k elements with the largest values are retained in the output logits vector. In this way, only the most representative predictions are transmitted.

Given an input sample $x$, the client $n$ calculates the complete logit  vector $\mathbf{K}_n(x) = (K_{n,1}(x), K_{n,2}(x), \dots, K_{n,c}(x))$, where $c$ is the total number of dimensions. The client then identifies the index set $I_{n,k}(x)$ corresponding to the largest k logit, which can be expressed as
\begin{equation}
    I_{n,k}(x)
= \bigl(K_{n,(1)}(x),\,K_{n,(2)}(x),\,\dots,\,K_{n,(k)}(x)\bigr),
\end{equation}
where \(K_{n,(1)}(x)\ge K_{n,(2)}(x)\ge \cdots \ge K_{n,(c)}(x)\) are the logits value in descending order. 
Therefore, each client $n$ transmits a sparse logits vector $\Bar K_n(x)$ for each sample $x$,  retaining only the logits associated with the top‑k most confident predictions. The sparse logits vector $\tilde{K}_{n,c}(x)$ of client $n$  is defined as

\begin{equation}
\tilde{K}_{n,c}(x) = K_{n,c}(x) \cdot \mathbb{I}[c \in I_{n,k}(x)],
\label{eq2}
\end{equation}
where $K_{n,c}(x)$ denotes the original logit value of dimension $c$. $I_{n,k}(x)$ denotes the set of Top-k    dimension indices for sample $x$ and $\mathbb{I}[\cdot]$ is the indicator function.

Before uploading, each client dynamically adjusts its Top-k value according to the proportion of channel resources ($\eta$) currently allocated to it. Then we assume that the communication link between a client and the server can be modeled as an Additive White Gaussian Noise (AWGN) channel. According to Shannon's theorem, the channel capacity $C$ (in bits per second) is given by
\begin{equation}
C = B \log_2(1 + \mathrm{SNR}),
\label{eq3}
\end{equation}
where $B$ is the channel bandwidth (Hz) and $\mathrm{SNR}$ is the signal-to-noise power ratio. $d$ denote the number of bits needed to encode one logit and its corresponding dimension index. Thus, the total transmission size for Top-k logits is $ k \times d$.

Further, let $T$ be the maximum transmission time (in seconds) allowed per communication round. Then the total bits that can be sent in time $T$ under fraction $\eta \in (0,1)$ of channel usage is $\eta \,C\,T.$ Based on each client channel condition, The maximum permissible Top-k   by $
k = \left\lfloor \frac{ \eta \times C \times T }{ d } \right\rfloor ,
$ where $\left\lfloor \cdot \right\rfloor$ denotes the floor operation.

During the uploading process, each client transmits sparse logits whose dimensionality is adapted to local channel conditions based on $k$. After receiving these sparse logits, the server aggregates them. The aggregated global logits representation can be used for subsequent distillation loss minimization.

Since each client uploads sparse logits, the missing entries must be handled during global aggregation. A common approach is to apply zero-padding on each logit vector to match the maximum length before averaging. However, this strategy has notable drawbacks, especially for LLMs (LLMs). Due to the high dimensionality of LLM outputs, zero-padding produces extremely sparse vectors, which weakens the quality of the aggregated results. 

Under Non-IID conditions, this issue becomes worse, as zero-padding further increases the differences between clients. Each client learns a unique output pattern based on its local data, which leads to biased and noisy logits. Aggregating logits from multiple clients helps smooth out this noise and produces more stable soft labels. However, the way sparse logits are combined plays a key role in determining the effectiveness of the distillation process.

We introduce an adaptive logit aggregation scheme that scales contributions according to each client’s transmitted dimension to alleviate these issues. We perform a dimension‐wise, sparsity‐aware aggregation of the Top-k logits.  Let $\scalebox{1}{$\bar{K}_n(x)$}$ be the sparse logit uploaded by client $n$ for dimension $c$ on sample $x$. 
Therefore, for each client $k$ and dimension $c$, define
$s_{n,c}= \lvert \tilde K_{n,c}(x)\rvert $.

Then, we collect these scores across all clients for dimension $c$, which can be expressed by $S[c]=\sum_{n=1}^{N} s_{n,c}.$ For each dimension \(c\), we compute the normalized contribution weight of each client is 
\begin{equation}
    w_{n,c}=\frac{s_{n,c}}{S[c]}.
\end{equation}

Finally, the global aggregated logit for dimension $c$ and sample $x$ is
\begin{equation}
K_{\mathrm{g},c}(x)
=
\sum_{n=1}^{N} w_{n,c}\;\tilde K_{n,c}(x).
\label{eq11}
\end{equation}

Only those clients that actually uploaded a non‐zero logit for dimension $c$ contribute to the sum. This adaptive, dimension‐wise normalization prevents zero‐padding bias and emphasizes high‐confidence predictions. At the end, eq~\eqref{eq11} preserves relative information density, mitigates sparsity, and yields more robust global updates compared with zero-padding, especially for high-capacity LLMs in heterogeneous data regimes.

\subsection{LoRA-Projection Alignment Distillation}
Conventional knowledge distillation aligns the student (client-side) and teacher (server-side) models by minimizing discrepancies between their output distributions (logits). however this output-level criterion is inherently limited: it fails to capture the rich intermediate latent representations learned by large-scale language-model teachers, thereby constraining the student’s ability to inherit the teacher’s internal knowledge structure.

To more effectively extract and transmit the teacher’s expressive internal knowledge, we propose an enhanced knowledge-distillation framework. The framework captures intermediate projection from the teacher’s LoRA and uses them as a complementary representation loss, optimized jointly with the conventional logits objective. Because LoRA projection are substantially lower-dimensional than complete logit vectors, they reduce communication overhead by roughly an order of magnitude while still carrying rich semantic structure. Experiments show that, with negligible extra distillation cost, the proposed scheme consistently improves the student model’s overall performance.


For each client $n$, let $\mathbf{x}\in\mathcal{D}_{p}$ denote an input vector from the public dataset, and let the associated LoRA adapter be parameterized by the low-rank factors $(\mathbf{A},\mathbf{B})$.  
The LoRA first projects $\mathbf{x}$ onto the low-dimensional subspace spanned by $\mathbf{A}$, producing the intermediate activation
\begin{equation}
h = \mathbf{A}\mathbf{x},
\qquad
h\in\mathbb{R}^{r}.
\label{eq7}
\end{equation}

Our overall distillation objective combines both the standard logits distillation loss and the activation distillation loss. This low-rank projection $h$ serves as the representation we extract for distillation. 

To compute the distillation loss on the public reference set
\(\mathcal D_{\text{p}}\), we first measure the
Kullback–Leibler divergence (KL) between the teacher’s and student’s
temperature-scaled probability distributions\cite{hinton2015distilling}, which can be described as
\begin{equation}
\mathcal{L}_{\text{logits}}
=
\frac{1}{|\mathcal D_{\text{p}}|}
\sum_{x\in\mathcal D_{\text{p}}}
\operatorname{KL}\!\Bigl(
\sigma\!\bigl(K_g(x)/T\bigr)
\;\Big\|\;
\sigma\!\bigl(\bar K_n(x)/T\bigr)
\Bigr),
\label{eq8}
\end{equation}
where {\(K_g(x)\)} and {\(\bar K_n(x)\)} are the global aggregated and client logits for example \(x\), \(T\) is the distillation temperature, and
\(\sigma(\cdot)\) denotes the softmax.

LoRA decomposes the original large weight matrix into two compact, low-rank factors. Consequently, the forward pass continues to produce a vector of logits. After applying the soft-max function, these logits naturally represent probability distributions. KL divergence is well-suited for measuring the difference between two probability distributions. Thus, we add the LoRA projection KL divergence $\mathcal{L}_{\text{h}} $ into the overall distillation objective using the same loss (from Eq. \ref{eq8}) as for the output logits. 

Therefore, our total distillation loss can be described as
\begin{equation}
\mathcal{L}_{\text{total}} = \mathcal{L}_{\text{logits}} + \lambda \cdot \mathcal{L}_{\text{h}},
\end{equation}
where $\lambda$ are weighting coefficients balancing the two loss terms. For the weight $\lambda$, we empirically determined the optimal value to achieve the best model performance. In the experiments of Section IV, we found that setting $\lambda$ between 0.03 and 0.5 yields the most favorable results.

It is worth noting that this distillation loss can be applied on both the client and the server sides. If the server-side model is a larger-scale language model with the same architecture, we can directly aggregate all intermediate projections $h$.

By incorporating intermediate activation distillation, our scheme allows the student model to better mimic the teacher model's internal feature processing, leading to improved performance compared to using only logits distillation.

Therefore, our overall scheme is shown in Algorithm \ref{alg1}.

\begin{algorithm}[ht]
\caption{Federated AdaLD scheme}
\label{alg1}
\begin{algorithmic}[1]
  \Require Global model $W_g$, user models $\{W_{u,n}\}_{n=1}^N$, public dataset $\mathcal{D}_p$, private datasets $\{\mathcal{D}_{u,n}\}_{n=1}^N$
  \State \textbf{Server:}  
  \State \quad Broadcast $\{K_g, h_g\}$ to all users
  \For{each user $n = 1 \dots N$ \textbf{in parallel}} 
    \State \quad $k_n, h_n \gets \mathrm{Infer}(W_{u,n}, \mathcal{D}_p)$
    \State \quad $L_{\text{logit}} \gets \mathrm{KL}(K_g \,\|\, k_n)$
    \State \quad $L_{\text{lora}}  \gets \mathrm{KL}(h_g \,\|\, h_n)$
    \State \quad Update $W_{u,n}$ with gradient of $ \mathcal{L}_{\text{total}} = \mathcal{L}_{\text{logits}} + \lambda \cdot \mathcal{L}_{\text{h}}$
    \State \quad $W_{u,n} \gets \mathrm{Train}(W_{u,n}, \mathcal{D}_{u,n})$
    \State \quad $ k_n,  h_n \gets \mathrm{Infer}(W_{u,n}, \mathcal{D}_p)$
    \State \quad $\hat k_i \gets \mathrm{TopK}(K, k_i)$ on real-time channel condition
    \State \quad Upload $(\hat k_n, h_n)$ to server
  \EndFor
  \State \textbf{Server:}
  \State \quad $\ K_s, h_s\ \gets \mathrm{Infer}(W_g, \mathcal{D}_p)$ 
  \State \quad Aggregate $\{\hat k_n\}$, $\{ h_n\}$ into $K_g, h_g$
  \State \quad Update $W_g$ by distilling $K_g, h_g$
\end{algorithmic}
\end{algorithm}
\vspace{-7pt}
\subsection{Communication Cost Analysis}

In traditional knowledge distillation, transferring model logits incurs significant communication overhead, as their size scales linearly with the output dimension and the number of samples, i.e., (number of samples * output dimension). In contrast, we propose extracting intermediate projection from inserted LoRA adapters, where each sample is associated with a low-dimensional vector of rank \( r \). Thus, the total output size is (number of samples + $r$) * output dimension. Notably, the total number of introduced LoRA projection is significantly smaller than the number of logits per sample, especially in large models with output dimensions in the hundreds or thousands. Despite their small size, these projection retain rich semantic information, enabling more efficient communication and knowledge transfer. Our experiments validate this insight.



\section {Numerical Results}

In our experiments, we utilize the GPT-2 \cite{radford2019language} series model as the primary architecture and conduct evaluations on the Banking77 dataset \cite{casanueva2020efficient}. On the client side, we deploy the GPT2-small model, while the server hosts a GPT2-large model. The experimental setup involves 50 clients and a single central server, with the dataset evenly partitioned among the 50 clients. We assume that the data distribution varies across clients, reflecting a Non-IID setting. A shared public dataset consisting of 2,000 samples is made available to all clients. In each communication round, a random subset of 10 clients is selected to participate in model training.


\begin{table}[ht]
  \centering
  \caption{}
  \label{t1}
  \scriptsize
  \setlength{\tabcolsep}{4pt} 
  \begin{tabular}{ll|ll}
    \toprule
    \textbf{Parameter}                  & \textbf{Value}       & \textbf{Parameter}                  & \textbf{Value}       \\
    \midrule
    Dataset                             & Banking77            & LoRA rank ($r$)                   & 8                    \\
    Samples per client                  & 2\,000               & Batch size                         & 32                   \\
    Total inquiries                     & 13\,083              & Learning rate                      & 0.001                \\
    Intent categories                   & 77                   & Weight decay                       & 0.001                \\
    LoRA $\alpha$                       & 32                   & LoRA dropout                       & 0.1                  \\
    Clients selected per round          & 10                   & Distillation temperature ($T$)     & 2.0                  \\
    Random seeds                        & 0,1,42,...           & LoRA distill.~weight ($\lambda$)   & 0.03                 \\
    \bottomrule
  \end{tabular}
  \vspace{-3mm}
\end{table}

To mimic realistic statistical heterogeneity across $n$ clients, we partition the dataset by class using a Dirichlet draw with concentration parameter $\gamma=0.5$. The detailed experimental setup is shown in Table \ref{t1}.

\begin{figure}[htbp]
\centering
\includegraphics[width=8cm]{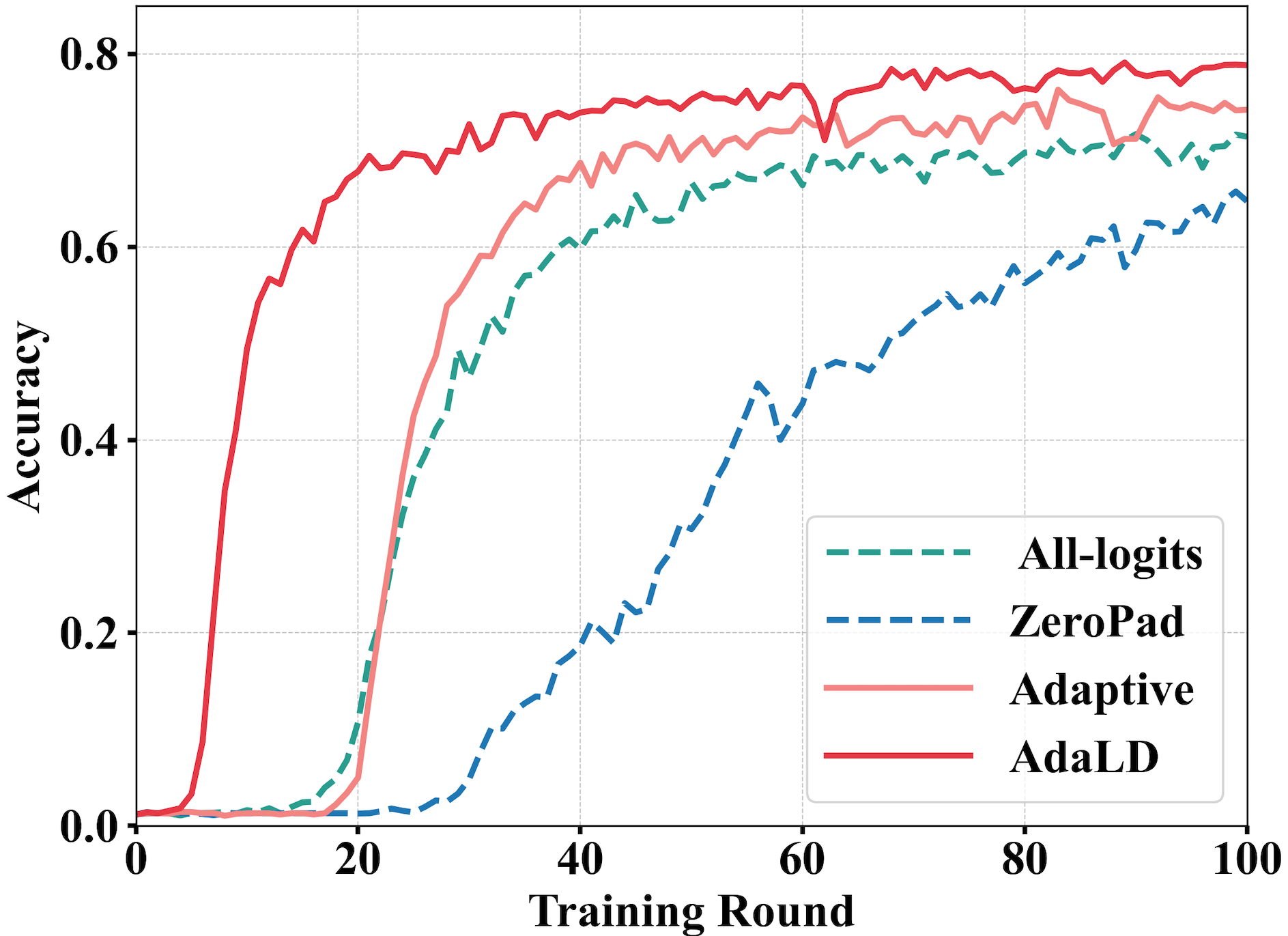}
\vspace{-7pt}
\caption{Performance for AdaLD and other schemes }
\label{fig2}
\end{figure}

In our experiments, we evaluated four methods. The first is AdaLD: our proposed Adaptive Logits Aggregation and LoRA Projection Alignment Distillation under limited communication resources. The second method is Adaptive: we use only adaptive logit aggregation. The third is ZeroPad: it applies traditional Zero-Padding aggregation, aligning inputs by zero padding. The fourth method is All-logits: it transmits all dimension logits without selection under sufficient communication resources, passing all predictions directly to the server. In the first three methods, due to limited communication bandwidth, the dimensionality of the logits transmitted by each user in each round will also vary. To ensure robustness, we report the average performance over three random seeds.

Fig. \ref{fig2} illustrates the accuracy trends of four different training strategies under a Non-IID data distribution. The figure depicts the performance of the server-side GPT2-large model, with the horizontal axis representing training rounds and the vertical axis representing accuracy. It can be observed that all methods exhibit slow accuracy growth in the initial training stages. As training progresses, accuracy gradually stabilizes. However, it is evident that our proposed scheme converges early and achieves the highest accuracy. Specifically, the All-logits method starts with relatively low accuracy but gradually improves over time, reaching around 0.7. This indicates that even though all logits were transmitted, the efficiency of the LLM was not significantly improved. The ZeroPad method shows very slow growth overall, with final accuracy not exceeding 0.6. This is largely due to the indiscriminate zero-padding, which introduces substantial distributional noise. In contrast, the Adaptive method is slightly slower than the All-logits method in early training but stabilizes at around 0.8 as training continues. The AdaLD scheme demonstrates outstanding early performance, with accuracy rising rapidly and ultimately reaching 0.85—clearly outperforming the other three methods. In summary, the AdaLD scheme achieves the best performance in this experiment, further validating the effectiveness of our proposed distillation techniques.
 \vspace{-2pt}
\begin{figure}[htbp]
\centering
\includegraphics[width=8cm]{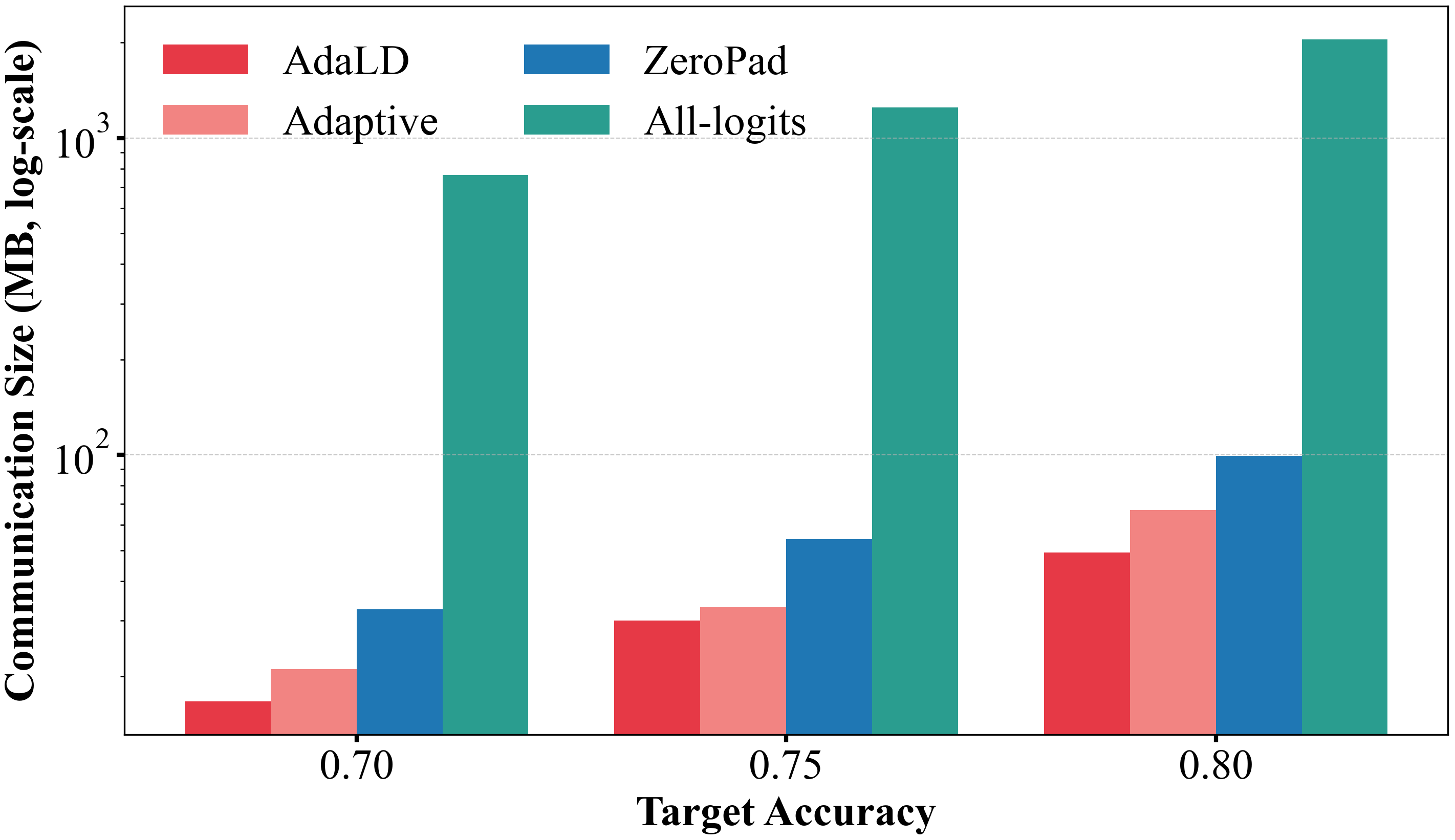}
 \vspace{-7pt}
\caption{Total communication cost comparison }
\label{fig3}

\end{figure}

Fig.~\ref{fig3} compares the total communication costs of four different strategies in achieving target accuracy thresholds (0.70, 0.75, and 0.79). Due to the large variance in communication under Non-IID settings, we report the results under the IID scenario for a fair comparison. The vertical axis denotes the total communication volume (in MB), while the horizontal axis represents the accuracy thresholds to be reached. We observe that the AdaLD scheme consistently achieves the target accuracy with the lowest communication overhead. At the 0.70 accuracy threshold, it requires only about 16.6 MB, substantially lower than All-logits (763.1 MB) and ZeroPad (32.5 MB). This trend holds across all thresholds, and at 0.79 accuracy, AdaLD's communication cost (49.1 MB) is significantly lower than that required by Adaptive (67.0 MB) and ZeroPad (99.5 MB). While the All-logits method delivers relatively high accuracy, it suffers from severe communication inefficiency, exceeding 2049.6 MB at the 0.79 threshold. This highlights the redundancy involved in transmitting the full set of logits, especially in large-vocabulary settings. Although the ZeroPad method is slightly more efficient than All-logits, it introduces a substantial amount of irrelevant information through zero-padding tokens. As a result, it requires more communication rounds to reach the desired accuracy, leading to a suboptimal efficiency-performance trade-off. In contrast, both Adaptive and AdaLD reach a better balance between communication volume and model accuracy. Notably, AdaLD achieves the lowest communication cost while maintaining high accuracy, indicating that the selective transmission of LoRA activation outputs contributes significantly more to model improvement than merely transmitting additional logits. This demonstrates that AdaLD effectively aggregates and distills informative features while avoiding excessive redundancy.

\section{Conclusion}

In this paper, we introduced a communication-aware federated distillation framework, leveraging a LoRA-projection-aligned loss combined with an adaptive weighted logits aggregation strategy. Specifically, by embedding LoRA projection alignment into the distillation objective and adaptively integrating logits across clients, our scheme significantly reduces inter-client communication by approximately 50\%, while concurrently enhancing model accuracy in both small and large scale language model scenarios. Extensive evaluations demonstrate that our method consistently surpasses three competitive baselines, validating the efficacy of exploiting LoRA-based projection insights and adaptive logit aggregation for achieving both lower bandwidth requirements and improved federated model performance.

\section{Acknowledgment}

This work was supported by UK Research and Innovation (UKRI) under the UK government’s Horizon Europe funding guarantee (grant number 10087666), as part of the European Commission-funded collaborative project MYRTUS, under the Smart Networks and Services Joint Undertaking (SNS JU) program (grant number 101135183).

\footnotesize
\bibliographystyle{IEEEtran}
\bibliography{conference_main}

\vspace{12pt}
\color{red}

\end{document}